\documentclass[10pt,twocolumn,letterpaper]{article}

\usepackage{iccv}
\usepackage{times}
\usepackage{epsfig}
\usepackage{graphicx}
\usepackage{amsmath}
\usepackage{amssymb}
\usepackage{multirow}
\usepackage[british,UKenglish,USenglish,english,american]{babel}
\usepackage{comment}
\usepackage[british,UKenglish,USenglish,american]{babel}
\usepackage{subcaption}
\usepackage{array}
\graphicspath{{figures/}}

\newcolumntype{x}[1]{>{\centering\arraybackslash\hspace{0pt}}p{#1}}

\usepackage[pagebackref=true,breaklinks=true,letterpaper=true,colorlinks,bookmarks=false]{hyperref}

\iccvfinalcopy 

\def\iccvPaperID{****} 

\ificcvfinal\fi
\begin{document}
\title{WordSup: Exploiting Word Annotations for Character based Text Detection}

\author{Han Hu$^1$\thanks{Equal contribution. This work is done when Han Hu is at IDL, Baidu Research.}  \quad Chengquan Zhang$^{2*}$ \quad Yuxuan Luo$^2$ \quad Yuzhuo Wang$^2$ \quad Junyu Han$^2$ \quad Errui Ding$^2$\\
Microsoft Research Asia$^1$ \quad IDL, Baidu Research$^2$\\
{\tt\small hanhu@microsoft.com \{zhangchengquan,luoyuxuan,wangyuzhuo,hanjunyu,dingerrui\}@baidu.com}
}

\maketitle

\begin{abstract}

Imagery texts are usually organized as a hierarchy of several visual elements, i.e. characters, words,
text lines and text blocks. Among these elements, character is the most basic one for various languages such as Western, Chinese, Japanese, mathematical expression and etc. It is natural and convenient to construct a common text detection engine based on character detectors. However, training character detectors requires a vast of location annotated characters, which are expensive to obtain. Actually, the existing real text datasets are mostly annotated in word or line level. To remedy this dilemma, we propose a weakly supervised framework that can utilize word annotations, either in tight quadrangles or the more loose bounding boxes, for character detector training. When applied in scene text detection, we are thus able to train a robust character detector by exploiting word annotations in the rich large-scale real scene text datasets, e.g. ICDAR15 \cite{karatzas2015icdar} and COCO-text \cite{veit2016cocotext}. The character detector acts as a key role in the pipeline of our text detection engine. It achieves the state-of-the-art performance on several challenging scene text detection benchmarks. We also demonstrate the flexibility of our pipeline by various scenarios, including deformed text detection and math expression recognition.
\end{abstract}

\section{Introduction}

Understanding optical texts has a long history dating back to the early twentieth century~\cite{schantz1982history}. For a long time, the attempts were made for texts of a few languages captured by very special devices, e.g. scanned English documents. With the growing popularity of smart phones, there have been increasing demands for reading texts of various languages captured under different scenarios.

\begin{figure}
\begin{center}
\begin{tabular}{c}
{\includegraphics[width=0.45\textwidth]{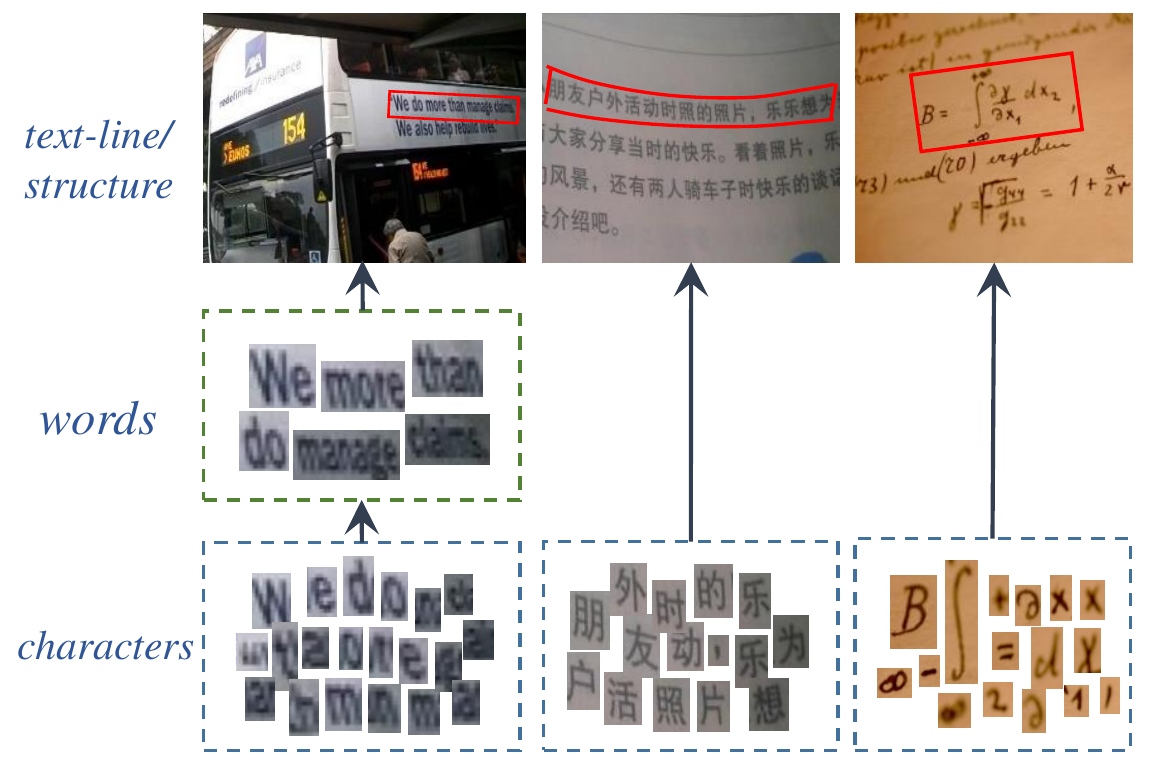}}
\end{tabular}
\end{center}
\vspace{-1em}
\caption{\small The visual hierarchies for various language texts under different scenarios. Different languages and scenarios may differ in hierarchy, but they are all formed by a basic element, \textit{character}.}
\vspace{-1em}
\label{fig:hierarchy_text}
\end{figure}

We are interested in developing a common text extraction engine for various languages and scenarios. The first step is to localize texts. It is not easy. Firstly, languages differ in organization structures. For an example, English texts include visual blank separation between words, while Chinese not. For another example, regular human language texts are organized sequentially, while math expressions are structural. Secondly, texts may differ in visual shapes and distortions according to the individual scenarios. Nevertheless, all optical texts share one common property that they are all formed by characters, as illustrated in Fig. \ref{fig:hierarchy_text}. It is natural and convenient that we base a common text detection framework on character detection.

When characters are localized, we can then determine the structure of texts in a bottom-up manner. The atomicity and universality of characters enable structure analysis for various languages and scenarios, e.g., oriented / deformed text lines and structural math expression recognition (see representative samples in Fig. \ref{fig:hierarchy_text}).

\begin{table}
\begin{center}
{\small
\begin{tabular}{|c|c|c|c|c|}
\hline
Dataset & \# im & \# word & {\scriptsize Real/Synth.} & Anno. \\
\hline\hline
ICDAR13~\cite{karatzas2013icdar} & 462 & 1,944 & Real & char\\
ICDAR15~\cite{karatzas2015icdar} & 1,500 & $\sim$12K & Real & word \\
SVT~\cite{wang2010word} & 350 & 725 & Real & word \\
COCO-Text~\cite{veit2016cocotext} & $\sim$63K & $\sim$174K & Real & word \\
{\small IIIT 5k-word~\cite{mishra2012scene}} & N.A. & 3000 & Real & word \\
\hline
Char90K~\cite{jaderberg2014synthetic} & N.A. & 9M & Synth. & char \\
{\scriptsize VGG SynthText}~\cite{Gupta16} & 800K & - & Synth. & char \\
\hline
\end{tabular}
}
\end{center}
\vspace{-1em}
\caption{\small Popular datasets and their properties. Nearly all median and large scale real datasets are annotated in word level.}
\label{tab:datasets}
\vspace{-1em}
\end{table}

The training of character detectors require a vast of location annotated characters. However, annotating character locations is very inconvenient and expensive, because the characters are small, easily gluing with each other and blurry. Actually, most existing large scale real text image datasets are labeled coarsely in word level, as illustrated in Table \ref{tab:datasets}.

\begin{figure}
\begin{center}
\begin{tabular}{c}
{\includegraphics[width=0.45\textwidth]{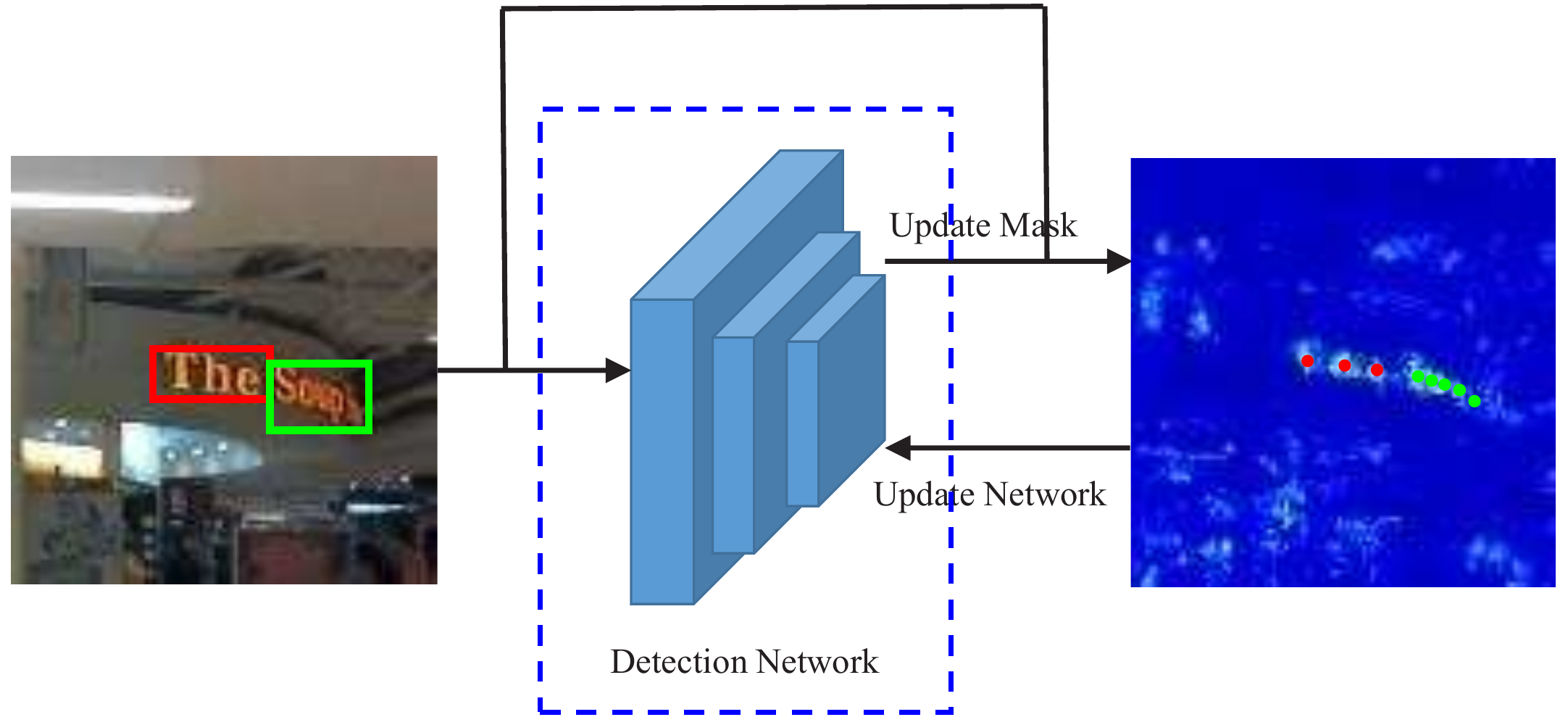}}
\end{tabular}
\vspace{-1em}
\end{center}
\caption{\small Illustration of our word supervision training approach for a character model. Two alternative steps are conducted: giving the current model, compute a response map which is then used together with word annotations to get a character center mask (red and green points); giving the character center mask, supervise the training of character model.}
\vspace{-1em}
\label{fig:weak_supervised}
\end{figure}

In this paper, we propose a weakly supervised learning framework to address the problem of lacking real character level annotations. It utilizes word annotations as supervision source to train the character detectors. Specifically, two alternative steps are iterated to gradually refine both the character center mask and the character model, as illustrated in Fig. \ref{fig:weak_supervised}. By applying this framework, we are able to train a robust character model by exploiting rich samples in several large scale challenging datasets, e.g. ICDAR15 \cite{karatzas2015icdar} and COCO-Text \cite{veit2016cocotext}.

The character model acts as a key module to our text detection pipeline. When applied to challenging scene texts, it achieves the state-of-the-art performance on several benchmarks, i.e. ICDAR13~\cite{karatzas2013icdar}, ICDAR15~\cite{karatzas2015icdar} and COCO-Text~\cite{veit2016cocotext}. It is also proved applicable on various scenarios, including deformed text line extraction and structural math expression recognition.

\subsection{Related Works}

There have been numerous approaches for text detection. According to the basic elements they rely on, the approaches can be roughly grouped into five categories:

\paragraph{Character based}
As mentioned earlier, character is a natural choice to build common detection engines. Nearly all existing character based methods rely on synthetic datasets for training \cite{tian2015text,TongHe16TIP,wang2011end, wang2012end,jaderberg2014deep,zhu2016text}, because of lacking character level annotated real data. However, synthetic data cannot have a full coverage of characters from various scenes, limiting the model's performance in representing challenging real scene texts. Actually, none of the current top methods for the popular ICDAR15 benchmark~\cite{karatzas2015icdar} are based on character detection. Recently, some sophisticated synthetic technologies \cite{Gupta16} have been invented that the synthetic text images look more ``real''. Nevertheless, real text images are still indispensable in training more robust character models, as we will show in our experiments.

Our pipeline is also character induced, but by incorporating a weakly supervised framework, we are able to exploit word annotations in several large scale real datasets to strengthen the character model. Using this model as a key to our pipeline, we achieve the state-of-the-art performance on several challenging scene text detection benchmarks. The pipeline is flexible for various scenarios such as deformed texts and structural math expressions.

\paragraph{Text Line based}

Text line based methods directly estimate line models. These methods are widely adopted in the field of document analysis~\cite{DBLP:conf/iccv/MengHSXP15}, where article layout provides strong priors. They are hard to be applied for non-document scenarios.

\paragraph{Word based}
A merit of these methods is that the modern object detection frameworks, such as faster RCNN \cite{DBLP:journals/corr/HuangYDY15} and SSD \cite{DBLP:conf/eccv/LiuAESRFB16}, can be conveniently adjusted \cite{jaderberg2016reading, Gupta16, liu2017deep, zhou2017east}. Yet, these methods are limited to languages which have word representation and visual separation between the words.

\paragraph{Component based}
Early component or word fragment based methods \cite{huang2013text,yin2014robust,huang2014robust,li2014characterness,yao2012detecting,yin2015multi,kang2014orientation,zhang2015symmetry} extract candidate text fragments by some manually designed features, e.g. MSER \cite{DBLP:conf/icip/ChenTSCGG11} and SWT \cite{epshtein2010detecting}, and then determine whether the fragments are real text or not. These methods once led some popular competitions for well focused texts, e.g. ICDAR13~\cite{karatzas2013icdar}. However, the performance of these methods heavily degrades when applied to more challenging scenarios such as ICDAR15~\cite{karatzas2015icdar} where texts are captured accidentally. In addition, as long as some texts are missed by the manually designed features, they would never be recalled in the subsequent steps.

Recently, some component based methods~\cite{zhang2016multi,yao2016scene,he2016accurate,DBLP:conf/eccv/TianHHH016,shi2017detecting} attempt to learn text components by CNN feature learning. The components are either representative pixels~\cite{zhang2016multi,yao2016scene,he2016accurate} or segment boxes~\cite{DBLP:conf/eccv/TianHHH016,shi2017detecting}. These methods can learn from word annotations. In addition, text component is also a basic visual element, which may also benefit a common text detection engine. Nevertheless, our method takes advantages over these methods in the following aspects: first, characters provide stronger cues, e.g., character scales and center locations, for the subsequent text structure analysis module; second, character is a semantic element, while component not. Thus our method is applicable to problems where direct character recognition is needed, e.g. match expression; third, our method can utilize loose word annotations for training, e.g. bounding box annotations in the COCO-Text dataset~\cite{veit2016cocotext}. This is because our method can refine character center labels during training. For the above component based methods, their noisy labels are fixed which may harm training.

\begin{figure}
\begin{center}
\begin{tabular}{l}
{\includegraphics[width=0.45\textwidth]{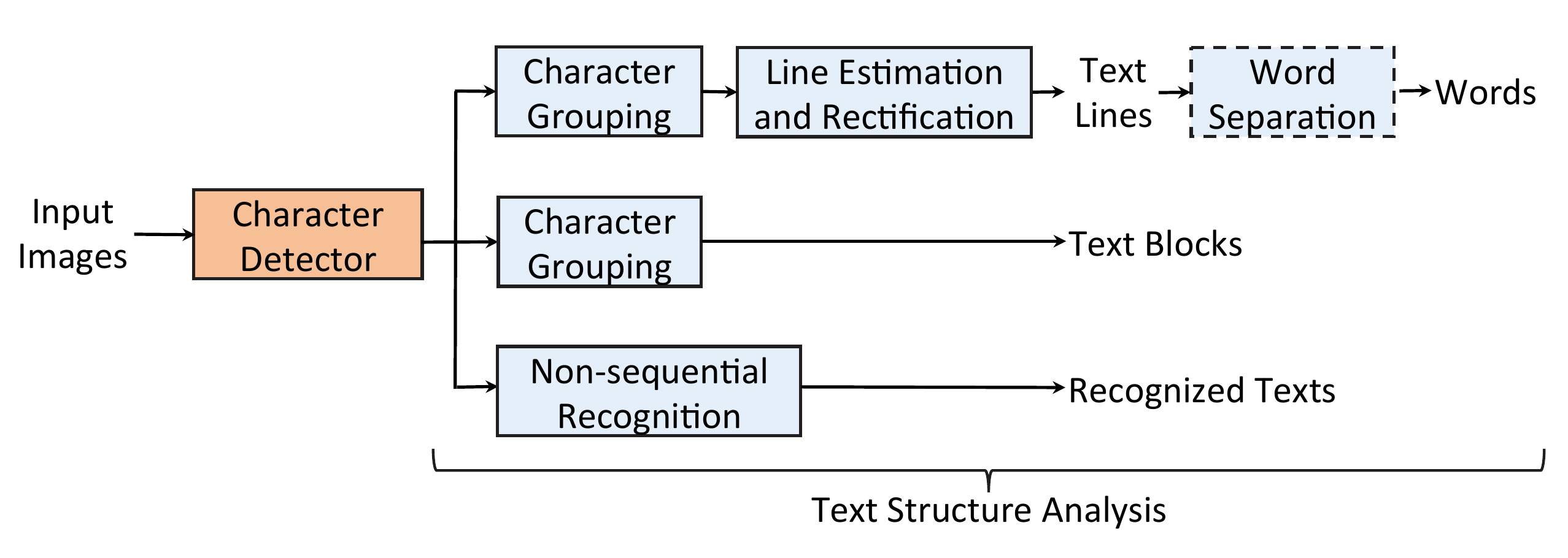}}
\end{tabular}
\vspace{-1em}
\end{center}
\caption{\small Our pipeline. There are two modules, character detector and text structure analysis. The pipeline is flexible for various scenarios ascribed to the atomicity and universality of characters.}
\vspace{-1em}
\label{fig:pipeline}
\end{figure}

\section {Our Approach}

\subsection {Pipeline}

The pipeline of our approach is illustrated in Fig. \ref{fig:pipeline}. Given an image, we first detect characters on it. This module is shared by various languages and scenarios. Its performance is crucial for the whole pipeline. Instead of using synthetic character data alone for training, we strengthen it by exploiting word annotations from real scene text datasets. The details of our basic character detector and the word supervision method are presented in Section \ref{sec:basic_network} and Section \ref{sec:weaksupervised}, respectively.

Then the detected characters are fed to a text structure analysis module, which is application dependent. We handle several typical scenarios. First is the sequential line, a widely used text structure. We propose a unified method to extract all of the horizontal, orientated and deformed text lines. English text lines are optionally separated into words for word based text recognition methods. Math expression recognition is another scenario, where characters are non-sequential. We first recognize all the detected characters and then
recover structures connecting characters/symbols \cite{He2017Context}. Details for text structure analysis are presented in Section~\ref{sec:text_ana}.

\subsection {Basic Character Detector}
\label{sec:basic_network}

The fully convolution neural network is adopted, which has seen good performance on general object detection, e.g., SSD~\cite{DBLP:conf/eccv/LiuAESRFB16} and DenseBox~\cite{DBLP:journals/corr/HuangYDY15}. Nevertheless, to be applied for characters, several factors need to be taken into account. First, characters may vary a lot in size on different images. Some characters may be very small, e.g., $10\times 10$ pixels in an 1M pixel image. Second, texts may appear in very different scenarios, such as captured documents, street scenes, advertising posters and etc, which makes the backgrounds distribute in a large space.

\begin{figure}
\begin{center}
\begin{tabular}{c}
{\includegraphics[width=0.45\textwidth]{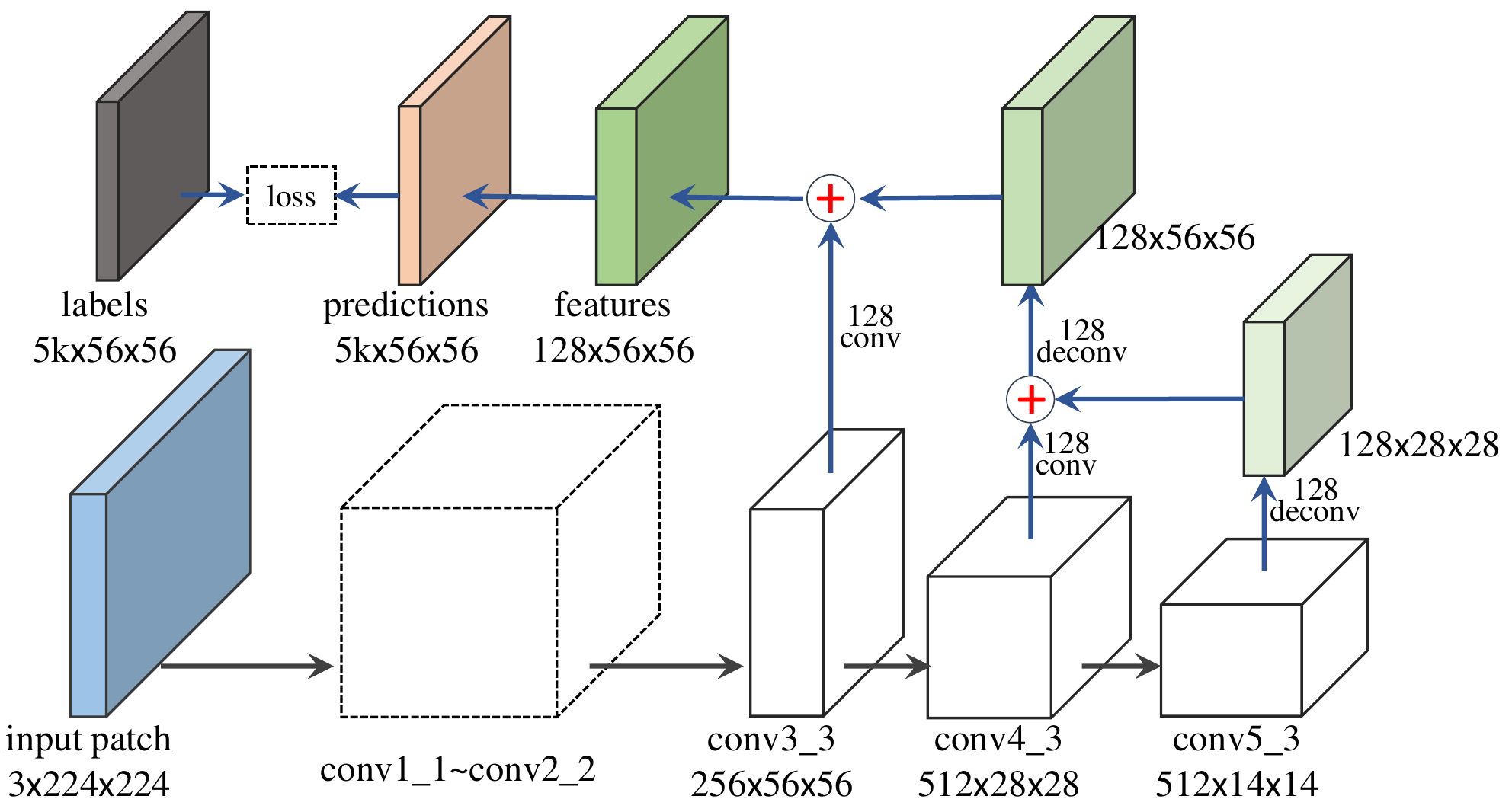}}
\end{tabular}
\vspace{-1em}
\end{center}
\caption{\small Our basic detection network. The network inherits from the VGG16 network model~\cite{DBLP:journals/corr/SimonyanZ14a}.}
\label{fig:basicDetection}
\vspace{-1em}
\end{figure}

To cope with the character size problem, we use feature maps with higher resolution to generate character responses. Specifically, it is $1/4$ of the original image, other than $1/16$ or $1/32$ used in general object detection~\cite{DBLP:conf/nips/RenHGS15,DBLP:conf/eccv/LiuAESRFB16}. Cues from deeper stages with coarser resolutions are merged for better representation power. We adopt the method in FPN~\cite{lin2016feature} for such purpose, which uses an \textit{eltsum} layer to merge features from two stages with $2\times$ resolution difference. It requires less parameters than other methods, e.g., \cite{long2015fully, ronneberger2015u, kong2016hypernet}, for producing the same number of feature maps. See Fig. \ref{fig:basicDetection} as an illustration. The network inherits from the VGG16 network model~\cite{DBLP:journals/corr/SimonyanZ14a}. \textit{conv5} features are $2\times$ up-sampled by deconvolution and merged with \textit{conv4} features by an \textit{eltsum} layer. The \textit{eltsum}ed \textit{conv5}-\textit{conv4} features are merged with \textit{conv3} features in the same way. The produced feature maps are used for both text/non-text classification and bounding box regression. $5k = (1+4)k$ score maps are generated, with $1$ for text/non-text classification, $4$ for bounding box regression, and $k$ indicating the number of anchors. We use $k=3$ anchors, representing characters with diagonal lengths of $24$ pixels, $16$ pixels and $12$ pixels (on the $224\times 224$ input patch), respectively. The characters with diagonal lengths of $0.7 \sim 1.4$ against the anchor's are regarded positive.

To ease the background variation problem, we adopt a two-level hard negative example mining approach for training. First is online hard negative mining~\cite{DBLP:conf/eccv/LiuAESRFB16}. All positives are used for loss computation. For negatives, only top scored ones are used in loss computation that the ratio between negatives and positives is at most $3:1$. Second is hard patch mining. During training, we test all training images every $10k$ iterations to find false positives (using the current character model). These false positives will be more likely sampled in the successive mini-batch sampling procedure.

\paragraph{Training} $32$ $224\times224$ patches are randomly cropped from training images to form a mini-batch. $50\%$ of the patches include characters. These positive patches are cropped from training images according to a randomly selected character and anchor, with some degree of translation/scale perturbation. The other $50\%$ are randomly cropped but with no texts. After $10k$ iterations, we start to apply the hard patch mining procedure. For negative patches, half training patches will be hard ones which should include the current detected false positives.

\paragraph{Inference} We conduct multi-scale test for an image. The used scales are $2^{\{0,-1,-2,-3,-4,-5\}}$, respectively. Since only down-sampling scales are involved, the computation overhead is afforded, at about $1.5$ times, compared to single-scale test. NMS with IoU threshold of $0.5$ is conducted to produce the final characters. It should be noted that multi-scale testing is indispensable for our basic detector, since we use anchors with only $3$ scales. Exploring more efficient basic detector without the need for multi-scale testing will be our future work.

\subsection {Learning From Word Annotations}
\label{sec:weaksupervised}

\paragraph {Overview}
As illustrated in Table \ref{tab:datasets}, most real text image datasets are annotated in word level, i.e. ICDAR15 and COCO-Text. Each word is attached with a quadrangle (e.g. ICDAR15) or a bounding box (e.g. COCO-Text) which tightly surrounds it, as well as a word category. In this paper, we suppose at least the bounding box of each word is available. If further a quadrangle or the word category is given, we use them to strengthen our word supervising procedure.

Our approach is inspired by~\cite{dai2015boxsup} which successfully learns object segments from bounding box annotations. It is illustrated as Fig. \ref{fig:weak_supervised}. Two alternative steps are conducted: given a character model, automatically generate the character mask according to a word annotation; given a character mask, update the character network. These two steps are alternative in each network iteration. During the training, the character masks and the network are both gradually improved.

It is worth noting that the above procedure is only involved in network training. The inference is the same as in Section~\ref{sec:basic_network}.

\begin{figure*}
\begin{center}
\begin{tabular}{c}
{\includegraphics[width=.95\textwidth]{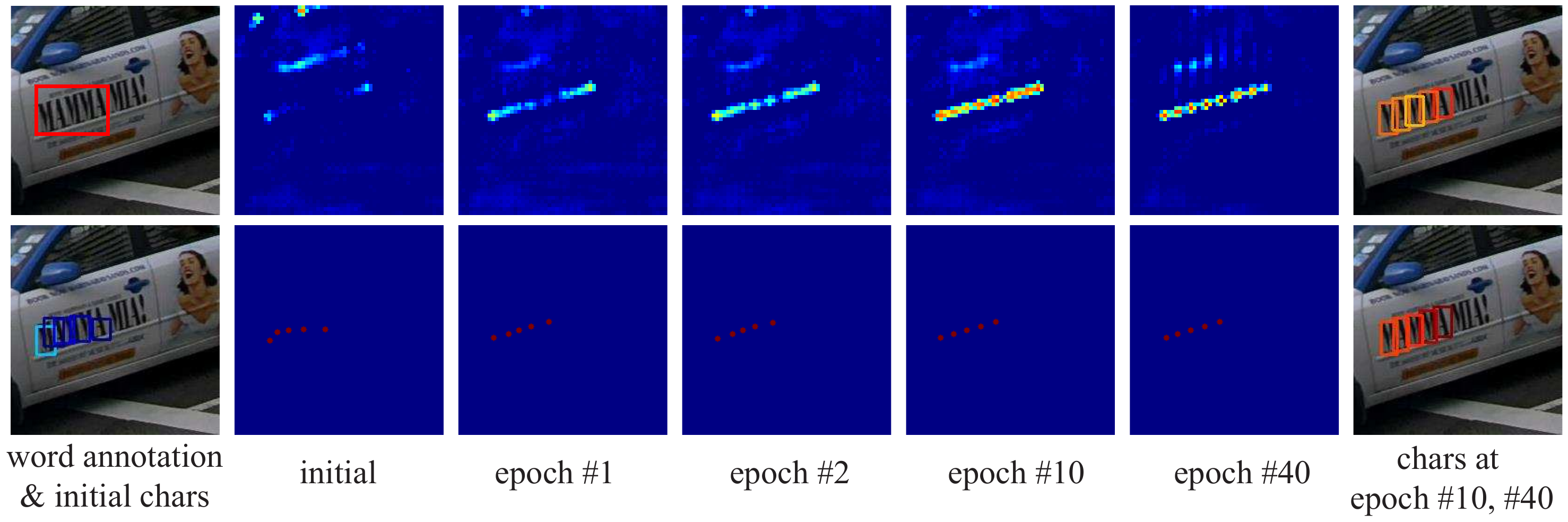}}
\end{tabular}
\vspace{-1em}
\end{center}
\caption{\small Updated character responses and the corresponding character masks during word supervised training on ICDAR15 datasets. The initial model in the second column is trained by $5k$ warmup iterations on synthetic data alone. The $3\sim 6^{th}$ columns are responses during the word supervised training, where the epoch number means for ICDAR15 datasets. For illustration, we use bounding box annotations rather than the original quadrangles in training. Both the responses and character candidates are colored by their scores (indexed by \textit{colormap} in Matlab).}
\label{fig:weak_supervised_mask}
\vspace{-1em}
\end{figure*}

\paragraph {Character Mask Generation}

During forward and backward of each mini-batch, the first step is to generate character masks using the current character model and word annotations, as illustrated in Fig. \ref{fig:weak_supervised_mask} (bottom). First, we make forward using the current character model and get a set of candidate characters inside the annotated word bounding box. We select real characters from these candidate characters by maximizing score,
\begin{equation}
\label{equ:sec3:score}
\begin{array}{*{20}{l}}
s & = w \cdot s_1 + (1-w) \cdot s_2 \\
& = w \cdot \frac{\text{area}(B_{\text{chars}})}{\text{area}(B_{\text{anno}})} + (1-w) \cdot (1-\frac{\lambda_2}{\lambda_1}),
\end{array}
\end{equation}
where $B_{\text{chars}}$ represents the bounding box of selected characters; $B_{\text{anno}}$ represents the annotated word bounding box; $\text{area}(\cdot)$ denotes the area operator; $\lambda_1$ and $\lambda_2$ are the largest and second largest eigenvalues of a covariance matrix $C$, computed by center coordinates of selected characters; $w$ is a weight balancing the two score terms. We find the learning is insensitive to the choice of $w$, and it is set as $0.5$ by default. The first term of Eq. (\ref{equ:sec3:score}), $s_1$, favors larger coverage of selected characters to the annotated word bounding box, while the second one, $s_2$, prefers that all characters locate on a straight line.

We use a similar approach as in \cite{DBLP:journals/pr/YinL09} to approximately maximize Eq. (\ref{equ:sec3:score}). Firstly, a maximum spanning tree \cite{DBLP:books/aw/AhoHU83}, $M$, is constructed from the character graph $G$, which is built by the $k$-nn of all candidate characters with pair weights defined by their spatial distances and the current text/non-text scores,
\begin{equation}
\label{equ:sec3:dist}
w_{ij} = exp(-\frac{d(i,j)}{\overline{d}}) \cdot (t_i+t_j),
\end{equation}
where $\overline{d}$ is the average of all distances between $k$-nn nodes; $t_i$ denotes the current text/non-text score of candidate $i$. It is obvious that eliminating an edge in $M$ equals to partitioning the characters into two groups. For each partitioning, we choose the group with larger score according to (\ref{equ:sec3:score}), and run the partitioning procedure greedily and recursively until the score (\ref{equ:sec3:score}) does not rise.

When a tight quadrangle or character number is given, we can further improve generated character mask: for the former, replacing computation of $s_1$ in Eq. (\ref{equ:sec3:score}) by area ratio of quadrangles; for the latter, adding a term to Eq. (\ref{equ:sec3:score}) that the mask prefers equal character number compared to ground truth.

\paragraph {Character Network Updating}

The generated character mask can be used as ground truth to supervise network training. We define a loss $\widetilde{\mathcal{L}}$ such that more reliable masks contribute more, as,
\begin{equation}
\label{equ:sec3:supervising}
\widetilde{\mathcal{L} } = s \mathcal{L},
\end{equation}
where $\mathcal{L}$ represents a combination of the confidence loss and localization loss commonly used in modern object detection frameworks \cite{DBLP:conf/nips/RenHGS15,DBLP:conf/eccv/LiuAESRFB16}; $s$ is the score computed by Eq. (\ref{equ:sec3:score}).

Fig. \ref{fig:weak_supervised_mask} shows the gradually updated character masks during training. During training, the character model is gradually improved.

\subsection{Text Structure Analysis}
\label{sec:text_ana}

Given characters extracted by the methods in Section~\ref{sec:basic_network} and ~\ref{sec:weaksupervised}, we conduct text structure analysis for various scenarios, e.g., text lines, words, text blocks, math expressions, and etc. Fig.~\ref{fig:pipeline} illustrates our text structure analysis methods for these typical text structures. For text line based applications, we propose a method which can handle arbitrarily deformed lines. The first step is to group characters. Then a line model is estimated to describe the line. With the model, we rectify the text line, which is usually required by modern sequential text recognition systems. Optionally, we separate lines into words. This is not necessary, but enables word based text recognition methods.

Characters can also be employed for text block extraction, e.g., document layout analysis~\cite{mao2003document}, and non-sequential text recognition, e.g., math expression recognition~\cite{hecontext}.

In the following, we briefly describe techniques used for extracting text lines, which are frequently used in our experiments. More details can be found in appendix.

\paragraph{Character Grouping}
We adapt the method in \cite{tian2015text} to group characters into text lines or blocks. Given characters with score larger than a threshold, \cite{tian2015text} first builds a $k$-nn graph with each node denoting a character candidate. Unary and pairwise costs are defined on the graph to achieve clustering. The unary costs model relations between characters and the text category, e.g. character scores. The pairwise costs model relations between two characters, e.g. spatial and scale distances. A greedy min-cost flow algorithm is then conducted to obtain all character groups (see \cite{tian2015text} for details).

The method in \cite{tian2015text} is designed for horizontal text lines only. To be applied in oriented and deformed text lines, we introduce a higher-order cost which models relations among three characters. To reserve the efficiency of pairwise graph, we use character pairs instead of characters as graph nodes. The character pairs are spatially close characters with high scores and small spatial/scale distances. Then the unary and pairwise costs in the old graph can be modeled as unary costs in the new graph, while the higher order costs, e.g. angle distance, can be modeled as pairwise costs in the new graph. The same as in \cite{tian2015text}, we then conduct a greedy min-cost flow algorithm on the new graph to achieve character grouping. It can handle oriented and deformed text lines, ascribed to the introduction of higher-order costs.

\paragraph{Line Model Estimation and Rectification}
For each character group, we fit three text line models with increasing complexity. First is \textit{0-order model}. Text lines are either horizontal or vertical. Second is \textit{1-order model}. Text lines can be arbitrarily orientated. Last is a \textit{piecewise linear model}, where a restricted polygon is adopted to represent a text line.

A model selection approach is conducted to choose a model with best balance between fitting accuracy and model complexity. Given the estimated model, we rectify the text line using thin plate spline (TPS)~\cite{bookstein1989principal} method, where the vertexes of the text line model are used as control points.

\paragraph{Word partition}
Some text recognition systems can process only word inputs. To enable usage of such systems, we optionally separate text lines into words. An LSTM~\cite{gers2000learning} based word blank detection method is applied on the rectified text line. Words are separated accordingly.

\section {Experiments}

In this section, we first do ablation studies on synthetic data where character level annotations are provided. Both our basic detector and the word supervision approach are evaluated. Then we apply our character induced text detection pipeline on scene text benchmarks. Finally, we show its applications to various scenarios.

\subsection {Datasets and Evaluation}

Four datasets are used in the experiments:
\begin{itemize}
  \item \textbf{VGG SynthText-part}. The VGG SynthText datasets~\cite{Gupta16} consist of 800,000 images, generated by a synthetic engine proposed in \cite{Gupta16}. The images have detailed character-level annotations. For experiment efficiency, we randomly select 50,000 images for training and 5,000 images for validation. This subset is referred to as \textit{VGG SynthText-part}.
  \item \textbf{ICDAR13}. The ICDAR13 datasets~\cite{karatzas2013icdar} are from the ICDAR 2013 Robust Reading Competition, with 229 natural images for training and 233 for testing. The texts are annotated with character-level bounding boxes, and they are mostly horizontal and well focused.
  \item \textbf{ICDAR15}. The ICDAR15 datasets~\cite{karatzas2013icdar} are from the ICDAR 2015 Robust Reading Competition, with $1000$ natural images for training and $500$ for testing. The images are acquired using Google Glass and the texts accidentally appear in the scene without user's prior intention. All the texts are annotated with word-level quadrangles.
  \item \textbf{COCO-Text}. The COCO-Text~\cite{veit2016cocotext} is a large scale dataset with 43,686 images for training and 20,000 for testing. The original images are from Microsoft COCO dataset.
\end{itemize}

The VGG SynthText-part is mainly used for ablation experiments. Both character-level and word-level evaluations are conducted by using the PASCAL VOC style criterion ($\geq 0.5$ Intersection-over-Union for a positive detection). For benchmark experiments on ICDAR13, ICDAR15 and COCO-Text, the evaluation protocols provided by the datasets themselves are adopted. Specifically, for ICDAR13 and ICDAR15, we use the online evaluation system provided with the datasets. For COCO-Text, the protocol provided by the dataset is used for evaluation.

\subsection {Implementation Details}
\label{sec:implementation_details}

The VGG16 model pretrained on the ILSVRC CLS-LOC dataset~\cite{DBLP:journals/ijcv/RussakovskyDSKS15} is adopted for all experiments.

Given different datasets, we train three character models. The first is trained by synthetic character data alone, i.e. 50k training images from VGG SynthText-part datasets. Second is trained on 1k ICDAR15 training images plus 50k VGG SynthText-part. $50\%$ are sampled from ICDAR15 and the others are sampled from VGG SynthText-part. The third is trained on COCO and VGG SynthText-part, with mini-batch also sampled half-half from the two datasets. These three models are dubbed as ``VGG16-synth'', ``VGG16-synth-icdar'' and ``VGG16-synth-coco'', respectively.

We use SGD with a mini-batch size of $64$ on 4 GPUs ($16$ per GPU). A total of 50k iterations are performed for all models. For the ``VGG16-synth'' model, 30k are at a learning rate of $0.001$, and the other 20k at $0.0001$. For other models, 5k iterations with VGG SynthText-part character supervision alone are first run for warming up. The learning rate is $0.001$ at this stage. Then 25k and 20k iterations are conducted using both character and word supervision at learning rates of $0.001$ and $0.0001$, respectively. The weight decay is set as $0.0005$ and the momentum as $0.9$.

For experiments on ICDAR13, ICDAR15 and COCO-text, the text line generation and word partition approaches introduced in Section~\ref{sec:text_ana} are applied to produce word localizations, which are required for evaluation of these benchmark datasets. For fair comparison, we tune hyperparameters of the line generation algorithm on a small fraction of training images, i.e. 50, for all character models.

\begin{figure}
\begin{center}
\begin{tabular}{c}
{\includegraphics[width=0.45\textwidth]{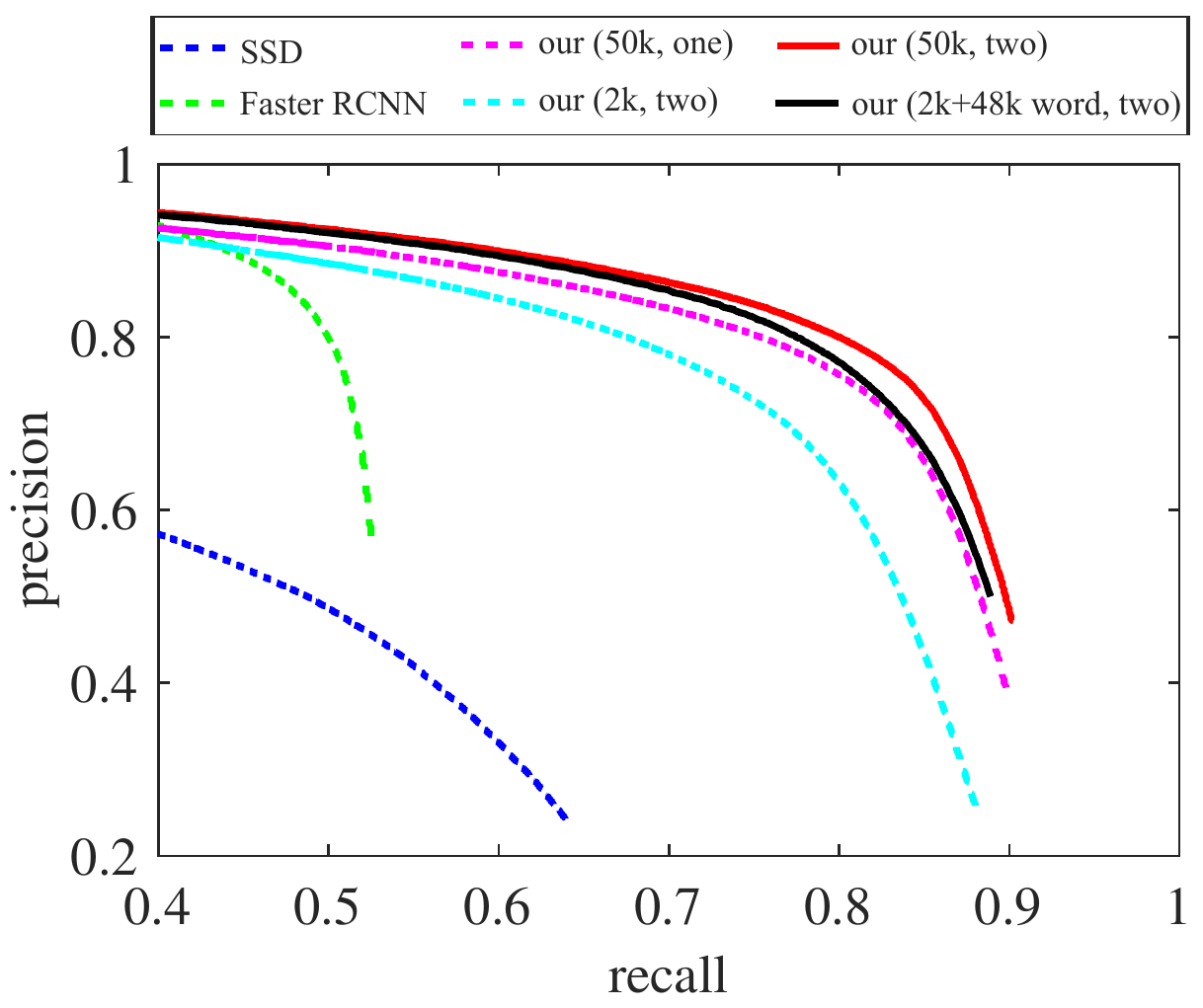}}
\end{tabular}
\vspace{-1em}
\end{center}
\caption{\small Character detection performance of our basic detection network, the faster RCNN and SSD methods on the VGG SynthText-part datasets. Four variants of our method are presented. The first term in brackets indicates the used supervision source. The second term indicates the used hard negative mining strategy, with ``one'' representing one-level hard negative mining and ``two'' representing two-level hard negative mining. }
\label{fig:comparison_ssd_fasterrcnn_our}
\vspace{-1em}
\end{figure}

\subsection {Experiments on Synthetic Data}
\label{sec:exp_abation_study}

The VGG SynthText-part datasets are used.

\vspace{-1em}
\paragraph{Evaluation of the Basic Character Detector}
We first compare the proposed basic detection network presented in Section \ref{sec:basic_network} with the state-of-the-art algorithms in the field of general object detection, e.g. faster RCNN \cite{DBLP:conf/nips/RenHGS15} and SSD \cite{DBLP:conf/eccv/LiuAESRFB16}. For faster RCNN and SSD, we directly use the codes provided by the authors.

Fig. \ref{fig:comparison_ssd_fasterrcnn_our} illustrates the precision-recalls of our basic network, faster RCNN and SSD on character detection, respectively. The main difference between our character network with the state-of-the-art general object detectors is that the feature maps used to produce character responses is finer than that of general object detectors ($1/4$ vs. $1/16$), while maintaining sufficient representation power by merging cues from deeper stages. The large gap between our basic network and general object detector demonstrates that reserving resolution is crucial for character detection. The two-level hard negative mining during training is also a plus that the second level hard patch mining can bring a moderate gain, as shown in Fig.~\ref{fig:comparison_ssd_fasterrcnn_our}.

\vspace{-1em}
\paragraph{Evaluation of Word Supervision Approach}
Three models are trained. The first is trained by randomly selected 2,000 images using character supervision. The second is trained using character supervision of all the 50k images. The third is trained using 2,000 character supervision images and 48,000 word supervision images. The training approaches are similar to those in \ref{sec:implementation_details}.

From Fig. \ref{fig:comparison_ssd_fasterrcnn_our}, it can be seen that the word-supervised model performs superior to 2k characters trained model and the performance degradation against the full 50k characters trained model is insignificant, demonstrating the effectiveness of our word supervision approach in exploiting weak word annotations for character model training.

\begin{figure*}
\begin{center}
\begin{tabular}{c}
{\includegraphics[width=0.95\textwidth]{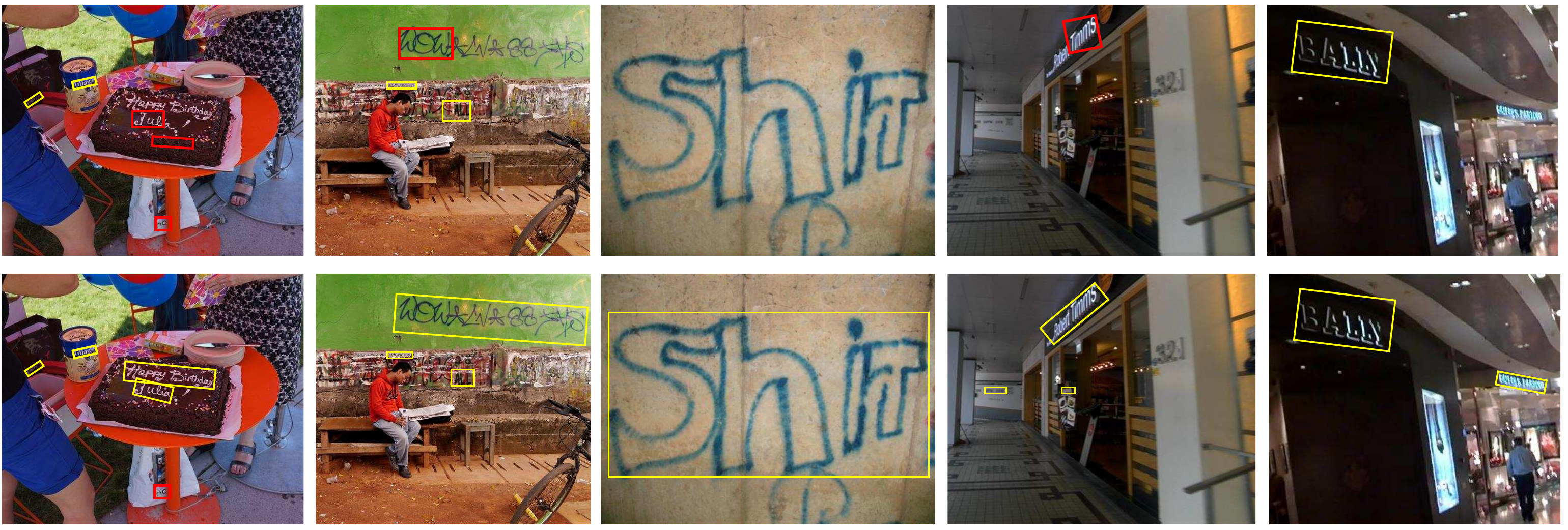}}
\end{tabular}
\vspace{-1em}
\end{center}
\caption{\small Sample qualitative results using the VGG16-synth model (top) and the models trained by word supervision (bottom) on the benchmark scene text datasets. Yellow and red rectangles illustrate the correctly and wrongly detected text lines, respectively.}
\label{fig:qualitative_res}
\vspace{-.5em}
\end{figure*}

\subsection {Experiments on Scene Text Benchmarks}
\label{sec:benchmark}

We apply our text detection approach on three real challenging scene text benchmarks: ICDAR13~\cite{karatzas2013icdar}, ICDAR15~\cite{karatzas2015icdar} and COCO-Text~\cite{veit2016cocotext}. These benchmarks are all based on word-level evaluation. Hence, the text line generation and word partition methods are involved. In the line model estimation step, we only use \textit{0-order} and \textit{1-order} models as nearly all text lines have up to orientation deformation.

\begin{table}
\begin{center}
\begin{tabular}{|c|c|c|c|}
\hline
Method & Recall & Precision & F-measure \\
\hline\hline
MCLAB-FCN \cite{zhang2016multi} & 79.65 & 88.40 &	83.80\\
Yao et al. \cite{yao2016scene} & 80.22 & 88.88 & 84.33 \\
Gupta et al.\cite{Gupta16} & 75.5 & 92.0 & 83.0 \\
Zhu et al. \cite{zhu2016text} & 81.64 & 93.40 & 87.13 \\
CTPN \cite{DBLP:conf/eccv/TianHHH016} & 82.98 & 92.98 & 87.69 \\
\hline
our ({\scriptsize VGG16-synth}) & 82.41 & 91.95 & 86.92 \\
our ({\scriptsize VGG16-synth-icdar}) & \textbf{87.53} & \textbf{93.34} & \textbf{90.34} \\
\hline
\end{tabular}
\vspace{-1em}
\end{center}
\caption{\small Performances of different methods on ICDAR13 using the DetEval criterion (\%).}
\label{tab:icdar2_res}
\vspace{-1em}
\end{table}

\begin{table}
\begin{center}
\begin{tabular}{|c|c|c|c|}
\hline
Method & Recall & Precision & F-measure \\
\hline\hline
MCLAB-FCN \cite{zhang2016multi} & 43.09 & 70.81 & 53.58 \\
CTPN \cite{DBLP:conf/eccv/TianHHH016} & 51.56 & 74.22 & 60.85 \\
Yao et al. \cite{yao2016scene} & 58.69 & 72.40 &64.77 \\
SCUT-DMPNet \cite{liu2017deep} & 68.22 & 73.23 & 70.64 \\
RRPN-2 \cite{Jianqi17} & 72.65 & 68.53 & 70.53 \\
\hline
our ({\scriptsize VGG16-synth}) & 64.37 & 74.79 & 69.18 \\
our ({\scriptsize VGG16-synth-icdar}) & \textbf{77.03} & \textbf{79.33} & \textbf{78.16} \\
\hline
\end{tabular}
\end{center}
\vspace{-1em}
\caption{\small Performances of different methods on ICDAR15 (\%).}
\vspace{-1em}
\label{tab:icdar4_res}
\end{table}

\begin{figure*}
\begin{center}
\begin{tabular}{c}
{\includegraphics[width=0.95\textwidth]{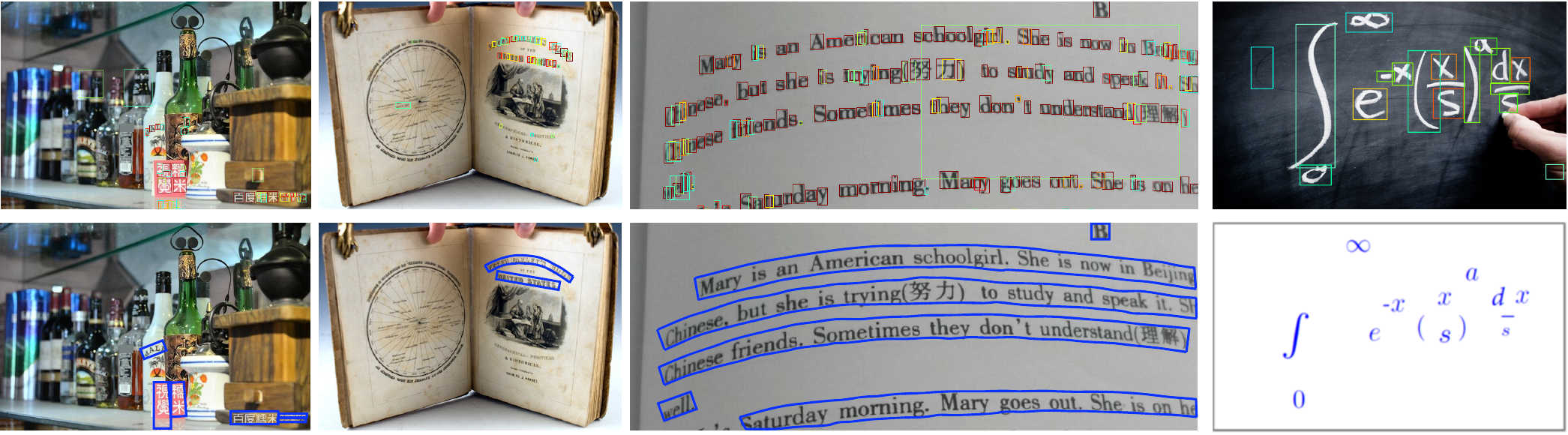}}
\end{tabular}
\vspace{-1em}
\end{center}
\caption{\small Applied to various scenarios. The top row shows detected characters, with colors indicating character scores (indexed by \textit{colormap} in Matlab). The bottom row shows results of structure analysis. }
\vspace{-1em}
\label{fig:various_app}
\end{figure*}

Table \ref{tab:icdar2_res},~\ref{tab:icdar4_res} and~\ref{tab:coco_res} show the performances of different methods on the ICDAR13, ICDAR15 and COCO-Text datasets. Our approach outperforms previous state-of-the-art methods by a large margin.

On ICDAR13, we achieve 90.34\% F-measure, which is 2.65\% higher than the second best one, i.e. CTPN \cite{DBLP:conf/eccv/TianHHH016}.

On the more challenging ICDAR15 datasets, images are more likely to suffer from blurry, perspective distortion, extreme illumination, and etc. Our best model achieves a f-measure of 78.16\%, with a large margin over the previous best method \cite{liu2017deep} (78.16\% vs. 70.64\%). Comparing our approach using different character models, VGG-synth-icdar performs much better than the VGG-synth model (78.16\% vs. 69.18\%). VGG-synth-icdar only adds \textit{1k} training image compared to the VGG-synth model (\textit{50k} training images). This indicates that the gain is from more \textit{real} data, other than more data.

On COCO, our best model achieves 30.9\%, 45.2\% and 36.8\% in recall, precision and F-measure, respectively. It takes over Yao's method by 3.5\% in total F-measure. VGG-synth-coco also performs much better than the VGG-synth model, demonstrating the introduction of real text images helps a lot in training better character models.

Fig. \ref{fig:qualitative_res} illustrates some detection samples from the ICDAR13, ICDAR15 and COCO-Text test images. By exploiting rich word annotations from real text image datasets, our model becomes more robust and can thus successfully detect various challenging texts, e.g. blurry, perspective distortion, handwritten/art fonts, extreme illumination and etc., which are hard to be synthesized using computers.

\vspace{-1em}
\paragraph{Computational Time}

For a $500 \times 500$ image, the character network takes about 500ms using an Nvidia Tesla K40 GPU. The text line generation and word partition procedures together take about 20ms using a 2GHz CPU.

\begin{table}
\begin{center}
\begin{tabular}{|c|c|c|c|}
\hline
Method & Recall & Precision & F-measure \\
\hline\hline
A~\cite{veit2016cocotext} & 23.3 & 83.78 & 36.48 \\
B~\cite{veit2016cocotext} & 10.7 & 89.73 & 19.14 \\
C~\cite{veit2016cocotext} & 4.7 & 18.56 & 7.47 \\
\hline
Yao et al.\cite{yao2016scene} & 27.1 & 43.2 & 33.3 \\
\hline
our ({\scriptsize VGG16-synth}) & 26.8 & 42.6 & 32.5 \\
our ({\scriptsize VGG16-synth-coco}) & \textbf{30.9} & \textbf{45.2} & \textbf{36.8} \\
\hline
\end{tabular}
\vspace{-1em}
\end{center}
\caption{\small Performance of different methods on the COCO-Text (\%). Notice that the annotations are obtained under the participation of method A, B and C. It is thus not fair to be compared with these methods. Yet, they are listed here for reference. }
\vspace{-1em}
\label{tab:coco_res}
\end{table}

\subsection{Applied to Various Scenarios}
\label{sec:various_scenarios}

We apply our pipeline to various challenging scenarios, including advertising images, deformed document texts and math expressions. A character model is trained by a privately collected text image datasets about these scenarios, consisting of $2k$ character-level annotated images and $30k$ line-level annotated images (only images with straight text lines are involved). The training approach is similar as in Section \ref{sec:implementation_details}. Text lines are generated by the approach in Section~\ref{sec:text_ana}. Fig.~\ref{fig:various_app} illustrates the character detection (top row) and text line generation (bottom row) results on some representative images. Our approach can handle text lines with various languages and extreme deformations. It is also worth noting that Chinese has a vast number of character classes, where some of them may not be seen by the $2k$ character-level annotated images. However, we empirically found that the initial model can still help recovering center masks of many unseen characters given only text line annotations. One possible reason is that the unseen chracters may share similar substructures or strokes with the characters seen by the initial model.

We also show application for math expression recognition (see the last column of Fig.~\ref{fig:various_app}). Math expressions are non-sequential, and hence sequential text recognition technique is not applicable. Given detected characters, we can recognize each of them, producing a set of math symbols.

\vspace{-.5em}
\section {Conclusion}
\vspace{-.5em}

Character based text detection methods are flexible to be applied in various scenarios. We present a weakly supervised approach to enable the use of real word-level annotated text images for training. We show the representation power of character models can be significantly strengthened. Extensive experiments demonstrate the effectiveness of our weakly supervised approach and the flexibility of our text detection pipeline.

\section{Appendix}

\paragraph{Character Grouping}

\begin{figure}
\begin{center}
\begin{tabular}{c}
{\includegraphics[width=0.45\textwidth]{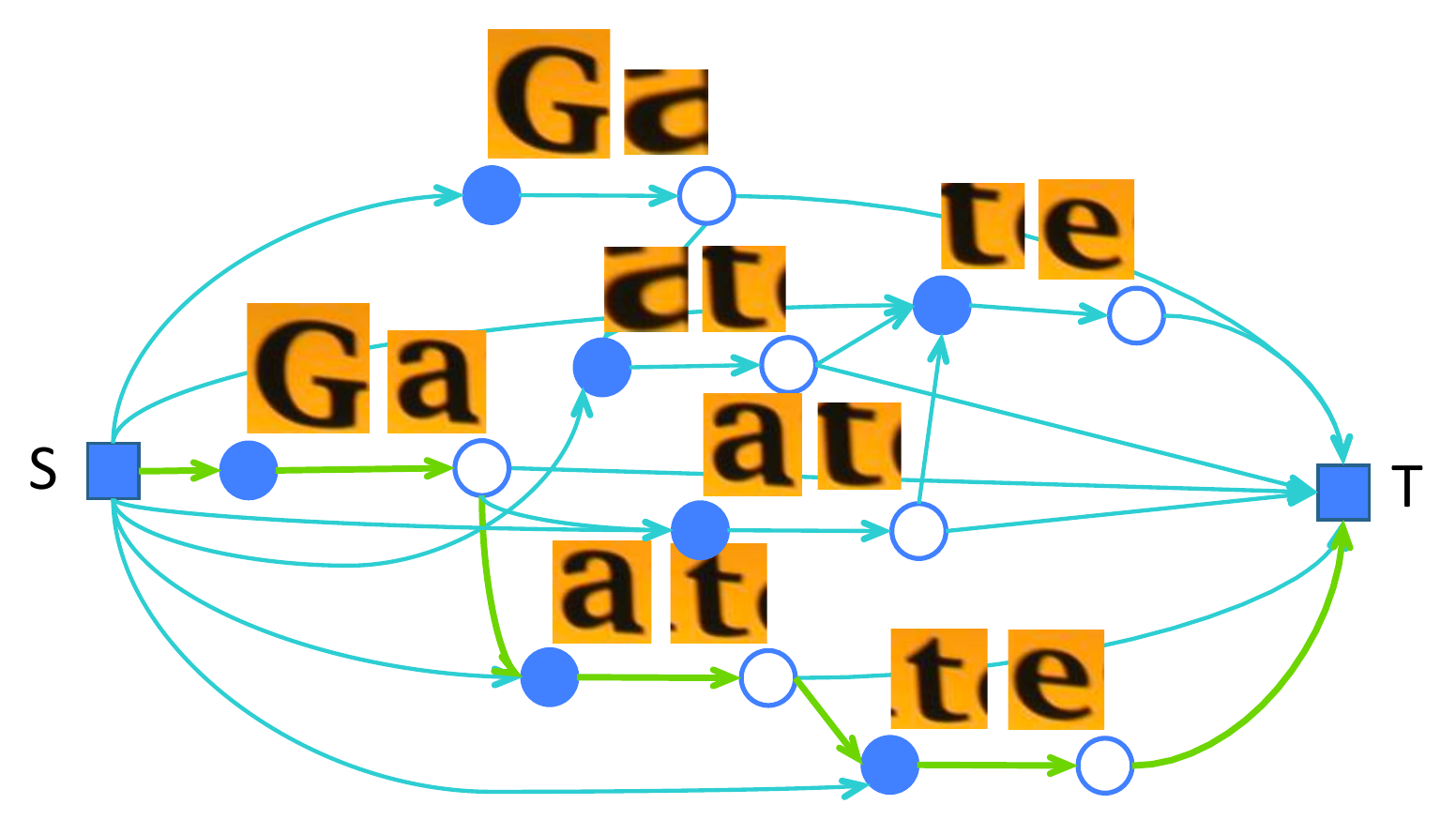}}
\end{tabular}
\end{center}
\caption{An illustration of the character grouping method. A min-cost flow algorithm is conducted n2on a built graph to achieve character grouping. Different from~\cite{tian2015text} which use characters as graph nodes, we use character pairs as graph nodes. The final text flow found by the greedy min-cost flow algorithm is illustrated by the green arrows.}
\label{fig:char_group}
\end{figure}

The method in~\cite{tian2015text} is adapted to group characters into text lines. The original method in~\cite{tian2015text} is designed for horizontal lines only. We adapt the method to handle multi-oriented text line generation. For this purpose, we propose to use character pairs instead of characters as graph nodes. Using character pairs, we can conveniently encode angle cues as the pairwise costs, and thus can handle multi-oriented text lines.

See Fig.~\ref{fig:char_group} as an illustration of our method. We compute the $k$ nearest neighbors of all characters according to the Euclidean distances of characters' center coordinates and diagonal length. The $k$ nearest neighbors form several character pairs and are set as the graph nodes. Graph edges are defined between character pairs which share one same character.

The flow costs in the min-cost flow graph are also redefined. The unary cost (associated with the graph node) is defined by the plus of average text/non-text score and center distances of two characters:
\begin{equation}
\label{eq:unary_cost}
C_{unary}= -\alpha \cdot (0.5 \cdot (p(n_l)+p(n_r))) + \beta \cdot d(n_l,n_r),
\end{equation}
where $n_l$ and $n_r$ denote the two characters of a character pair; $p(n_l)$ and $p(n_r)$ denote the text/non-text scores; and $d(n_l,n_r)$ is the Euclidean distance between two characters. The two terms in Eq.~(\ref{eq:unary_cost}) correspond to the unary and pairwise costs in the original paper~\cite{tian2015text}, respectively.

We define the pairwise costs (associated with the graph edge) by angle distances:
\begin{equation}
C_{pairwise}= cos(\theta(m, n)),
\end{equation}
where $\theta(m, n)$ is the angle distance between character pairs $m$ and $n$. The entry and exit costs are the same as in~\cite{tian2015text}. A greedy min-cost flow algorithm is conducted on the new graph to achieve character grouping.

\paragraph{Text Line Models}

A text line is represented by a set of center lines $\{\mathbf{l}_i\}_{i=1}^n$ and a height value $h$, where $\mathbf{l}_i = (a_i, b_i, c_i)$ represents a line of $a_i x + b_i y + c_i = 0$. For \textit{0-order} and \textit{1-order} models, one center line is estimated using all the center coordinates of characters ($n=1$) . For \textit{piecewise linear} model, $N$ line segments are estimated for each character by using its $k = \min(n, 11)$ nearest neighbors ($n=N$ with $N$ indicating the number of characters). The height value $h$ is set as $h = 2\max \limits_{\mathbf{p} \in P} {\min \limits _{i=1} ^{n}d(\mathbf{p}, \mathbf{l}_i)}$, where $P$ is the set of all character corner coordinates and $d(\mathbf{p}, \mathbf{l}_i)$ denotes the distance between a point $\mathbf{p}$ and a line $\mathbf{l}_i$.

We select the best line model $M$ by minimizing
\begin{equation}
M = \arg \min \limits_{m \in \{0, 1, \text{piece-wise}\}} h_m \cdot C_m,
\end{equation}
where $h_m$ denotes the estimated height value of model $m$, with smaller $h_m$ indicating better model fitting; $C_m$ denotes the model complexity penalty, set as $1.0$, $1.2$ and $1.4$ for \textit{0-order}, \textit{1-order} and \textit{piecewise linear} models, respectively.

\begin{figure}
\begin{center}
\begin{tabular}{c}
{\includegraphics[width=0.45\textwidth]{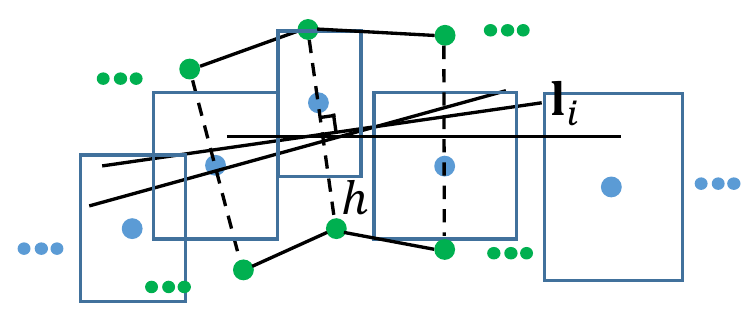}}
\end{tabular}
\end{center}
\caption{Illustration of text polygon computation. Character bounding boxes and their centers are plotted in light blue. Black solid lines represent the estimated center lines. For each character, two control points (green dots) are computed that they symmetrically locate on two sides of the center line and their connection goes through the character center point.}
\label{fig:text_polygon}
\end{figure}

\paragraph{Text Line Rectification}
The text lines are rectified to $H\times W$ strip images, with $H$ fixed as $32$. First, a closed polygon is computed to represent a text line according to the estimated center lines and height values, as illustrated in Fig.~\ref{fig:text_polygon}. The polygon has $2n$ vertexes / control points. Then, the $2n$ corresponding target control points on the rectified image are computed, which locate on the boundary. Finally, a thin plate spline (TPS)~\cite{bookstein1989principal} method is used to achieve the rectification.

\begin{figure}
\begin{center}
\begin{tabular}{c}
{\includegraphics[width=0.45\textwidth]{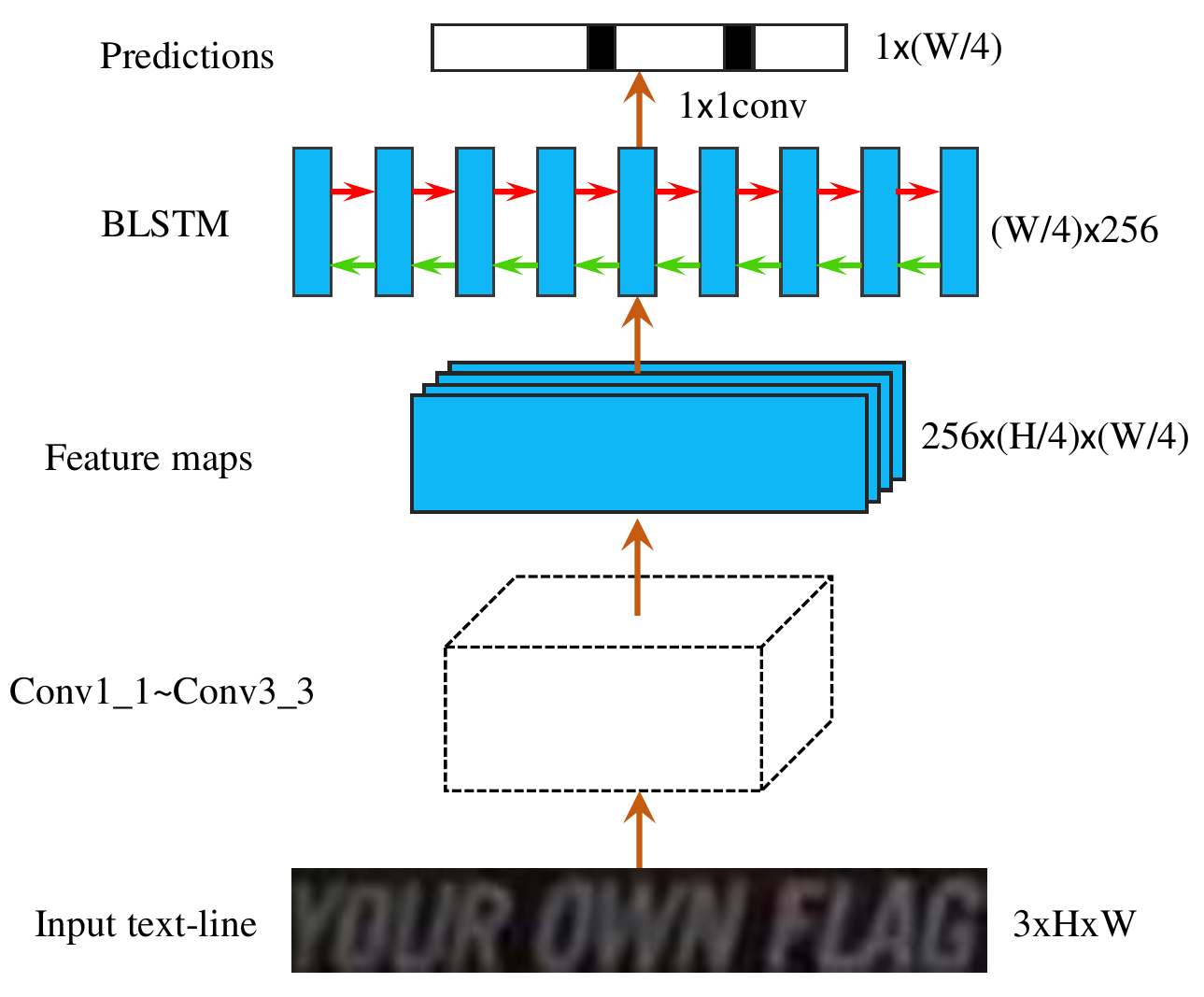}}
\end{tabular}
\end{center}
\caption{Word Partition Network.}
\label{fig:word_sep_net}
\end{figure}

\paragraph{Word Partition}
Word partition is not necessary for an OCR system. Yet, it enables the word based text recognition methods and is also required for the evaluation of several popular benchmarks, e.g., ICDAR13, ICDAR15 and COCO-Text. Hence, we optionally involve a word partitioning module in the text structure analysis stage. We propose a CNN-RNN approach for it, as illustrated in Fig.~\ref{fig:word_sep_net}. First, $7$ convolutional layers inheriting from VGG-16 net ($conv1\_1 \sim conv3\_3$) are applied on a rectified line image to produce feature maps with $1/4$ resolution of the original image. Then, a BLSTM layer~\cite{gers2000learning} along the \textit{horizontal} direction is involved to predict a sequence of labels, which indicates whether there is word separation or not at the place.

In training, 20k $32 \times 512$ rectified text line images are randomly generated from ICDAR15 and VGG-Synth-part datasets. For each sampled word, we automatically determine its connected words along the word text direction to form a text line. Padding and cropping are adopted for narrow and wide text lines, respectively, to make the rectified text line images with constant width $W = 512$. Shuffling, blur, noise and slight rotation($-5$ degree $\sim$ $5$ degree) are used for data augmentation. 40k iterations with mini-batch size of 32 are conducted. The learning rates are $0.001$ and $0.0001$ in the first $\frac{1}{2}$ and the last $\frac{1}{2}$ iterations, respectively.

For inference, we apply the network to rectified text lines with height of 32 pixels and width automatically determined by keeping the aspect ratio. The detected word separation positions are mapped back to produce word polygons, which are further transferred to required formats for benchmark evaluation.

{\small
\bibliographystyle{ieee}
\bibliography{egbib}
}

\end{document}